\documentclass[conference]{IEEEtran}
\IEEEoverridecommandlockouts
\usepackage{cite}
\usepackage{amsmath,amssymb,amsfonts,upgreek}
\usepackage{graphicx}
\usepackage{textcomp}
\newcommand{\figvspace}{\vspace{-0.65em}} 

\usepackage[english]{babel}
\usepackage[utf8]{inputenc}
\usepackage{algorithm}
\usepackage[noend]{algpseudocode} %
\def\eg{\emph{e.g.},}

\usepackage{colortbl} 
\usepackage{booktabs}
\usepackage[table,xcdraw]{xcolor}
\usepackage{wrapfig} 
\usepackage{multirow}

\def\BibTeX{{\rm B\kern-.05em{\sc i\kern-.025em b}\kern-.08em
    T\kern-.1667em\lower.7ex\hbox{E}\kern-.125emX}}
\begin{document}

\title{ReMoBot: Retrieval-Based Few-Shot Imitation Learning for Mobile Manipulation 
\\
\thanks{$^{*}$ Equal contribution}
\thanks{The authors are affiliated with School of Electrical Engineering, Aalto University, Espoo, Finland. {\tt\{firstname.surname\}@aalto.fi}.}
\thanks{The authors acknowledge the use of the MIDAS infrastructure of Aalto School of Electrical Engineering. V.\ Kyrki acknowledges the research environment provided by ELLIS Institute Finland.}
\thanks{This research was supported by the European Union’s Horizon Europe research and innovation program under grant agreement (No 101189836) for the XSCAVE project, and the Research Council of Finland for the MARL project (357301).}
}

\author{
Yuying Zhang$^{*}$,  Wenyan Yang$^{*}$, Francesco Verdoja, Ville Kyrki, Joni Pajarinen
}

\maketitle

\begin{abstract}

Imitation learning (IL) algorithms typically distill demonstrations into parametric policies to mimic expert behavior. However, with limited data and partial observability, such as in egocentric mobile manipulation, existing methods often struggle to generate accurate actions. To address these challenges, we propose ReMoBot, a few-shot, trajectory-conditioned imitation learning framework that directly \textit{Re}trieves information from demonstrations to solve \textit{Mo}bile manipulation tasks with ego-centric visual observations. Leveraging vision foundation models, ReMoBot identifies relevant expert demonstrations by combining state-level similarity, history-aware trajectory alignment, and action-sequence consistency to disambiguate perceptually similar observations. The agent then selects appropriate control commands based on these retrieved demonstrations in a fully training-free manner.

We evaluate ReMoBot on three mobile manipulation tasks using a Boston Dynamics Spot robot in both simulation and real-world settings. After benchmarking five approaches in simulation, we compare our method with two baselines trained directly on real-world data without sim-to-real transfer. With only 20 demonstrations per task, ReMoBot outperforms the baselines, achieving high success rates in Table Uncover (70\%) and Gap Cover (80\%), while also showing promising performance on the more challenging Curtain Open task in the real-world setting. Furthermore, ReMoBot generalizes across varying robot positions, object sizes, and material properties, highlighting its robustness in real-world deformable mobile manipulation. Additional details are available at: https://sites.google.com/view/remobot/home
\end{abstract}


\section{Introduction}
\begin{figure}
    \centering
    \centerline{\includegraphics[width=0.48\textwidth]{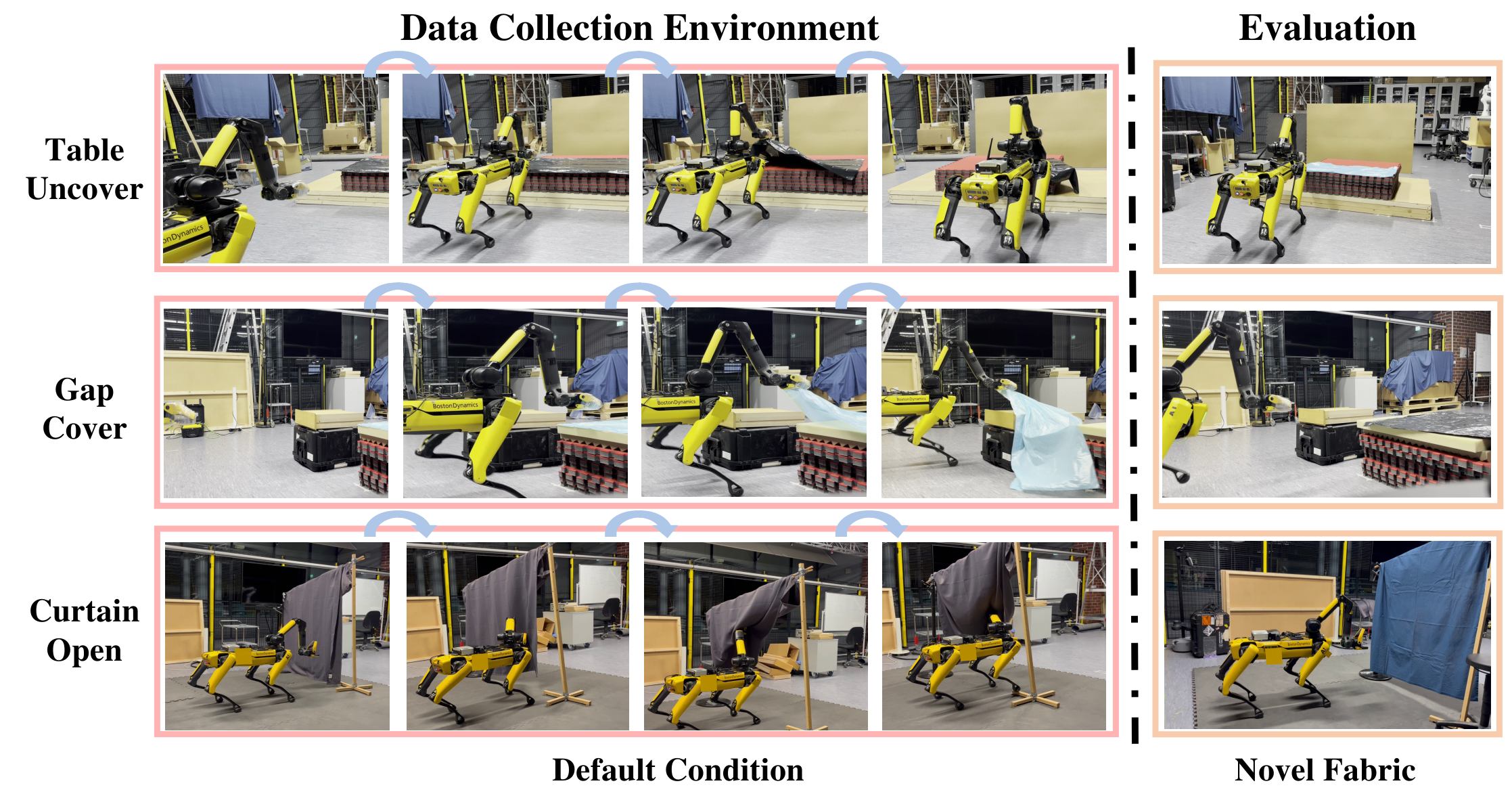}}
    \caption{\textbf{Three Deformable Mobile Manipulation tasks.} Table Uncover (top), Gap Cover (middle), and Curtain Open (bottom) are shown in both the data collection (left) and novel fabric evaluation (right) settings. 
     \figvspace{}
    }
\label{fig:task_vis}
\end{figure}

Learning mobile manipulation purely from egocentric visual inputs is challenging due to partial observability arising from a limited camera field of view and the complexity of the environment~\cite{ego_cv,ego_loco}. While reinforcement learning (RL) has shown promise in certain complex scenarios, it typically requires extensive exploration~\cite{rl_alg_book,visulmm_rl}, limiting its practicality for real-world applications without additional guidance. In contrast, imitation learning (IL) enables robots to efficiently acquire skills from expert demonstrations~\cite{mimicplay}. However, the performance of IL methods depends greatly on the quantity and diversity of demonstrations, and approaches such as behavior cloning are prone to compounding errors over long task horizons~\cite{BC_survey}. 

To mitigate these issues, retrieval-based imitation learning~\cite{dinobot,VINN} has been proposed, which leverages expert demonstrations directly rather than relying solely on parametric policies. Nevertheless, existing approaches primarily focus on static manipulation~\cite{dinobot} and retrieve actions based only on individual observations. These strategies are insufficient for egocentric mobile manipulation, as partial observability induces state ambiguity, where visually similar observations may arise from different underlying world states and thus necessitate distinct actions.

Motivated by these challenges, we investigate how far visual foundation models and retrieval-based imitation learning can be leveraged for egocentric mobile manipulation from limited demonstrations.
To this end, we introduce ReMoBot, a trajectory-conditioned non-parametric imitation framework for few-shot ego-centric mobile manipulation using only visual input. Rather than learning a parametric mapping from observations to actions, ReMoBot imitates demonstrated behaviors by retrieval over expert trajectories. Given the executed history, the next action is selected from the most distributionally aligned demonstration, enabling robust decision-making under partial observability without additional training. ReMoBot introduces two key contributions: (1) modeling mobile manipulation under partial observability as a trajectory-conditioned non-parametric policy, and (2) a history-aware retrieval mechanism that enforces trajectory-level similarity constraints, reducing observation aliasing and ensuring temporally consistent action selection. Through systematic evaluation on three deformable mobile manipulation tasks with a Boston Dynamics Spot robot, we examine the effectiveness of this approach in both simulation and real-world settings.

\begin{figure*}[thbp]
    \centering
        \centering
        \centerline{ \includegraphics[width=1\textwidth]{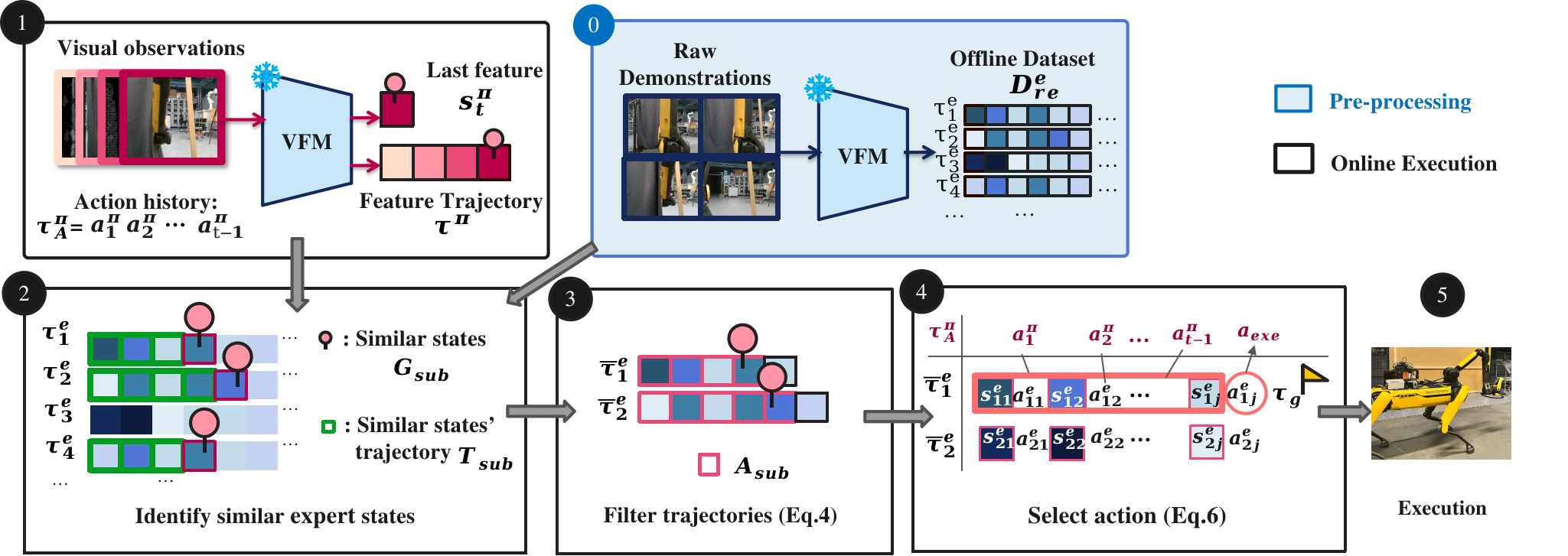}}

    \caption{\textbf{Overview of ReMoBot:} An offline dataset is first processed using a pre-trained visual foundation model (VFM) to build a retrieval dataset. During execution, ReMoBot encodes RGB observations into the same feature space to identify similar expert states. These candidates are then filtered based on trajectory similarity, and the agent selects the final action by matching action history, enabling efficient training-free task execution.
     \figvspace{}
}
\label{fig:overview}
\end{figure*}

\section{Related Work}
\label{sec:related_work}
\textbf{Vision-based mobile manipulation:}
Recent advances in vision-based mobile manipulation have improved generalization in real-world settings~\cite{leggedmm_survey}. However, egocentric viewpoints introduce persistent challenges, including occlusions, viewpoint shifts, and partial observability, which are further amplified when manipulating deformable objects with complex dynamics~\cite{cloth_folding1,cloth_unfolding1}. Existing policy learning approaches remain constrained by high exploration costs~\cite{open_door, mobilemanipulation_rigid}, large data requirements~\cite{mm_il_simdata,mmil_emma}, and insufficient mechanisms for handling incomplete observations~\cite{Mobile_ALOHA}. Therefore, these methods often fail to generalize from a few demonstrations in complex egocentric settings.

\textbf{Retrieval-based imitation learning:}
Retrieval-based imitation learning enables few-shot skill learning by reusing relevant demonstrations from offline datasets or memory libraries~\cite{strap, retrieval_behavior, handme, retri_memory}. Some methods directly execute retrieved behaviors via learned embeddings~\cite{il_replay_search, VINN}, flow-based alignment~\cite{flowretrieval}, affordance reasoning~\cite{roboabc}, or foundation models~\cite{dinobot}, while others integrate retrieval into policy optimization~\cite{retri_align, read, gsr_bc}. Despite their success, most retrieval-based approaches assume static viewpoints or rely on single-step state embeddings, which are not tailored to egocentric mobile manipulation under severe occlusions and dynamic viewpoints.

To compensate for the state representations, trajectory-level information has been incorporated via explicit alignment or distribution modeling~\cite{pomdp}. In retrieval settings, ~\cite{strap} employs Dynamic Time Warping(DTW) to stitch trajectory segments. However, DTW assumes strict temporal ordering and is sensitive to local misalignments, limiting its robustness in mobile manipulation. In contrast, Optimal Transport (OT)~\cite{wassertain} treats trajectories as distributions, providing robustness to noise and execution speed variations. Building on these insights, we propose a non-parametric retrieval mechanism guided by OT-based trajectory similarity, enabling training-free and data-efficient egocentric mobile manipulation.

\section{ReMoBot}
\label{sec:remobot}
We propose ReMoBot, a trajectory-conditioned non-parametric imitation policy for few-shot mobile manipulation. ReMoBot operates in three stages:  1) retrieval dataset generation, which extracts compact visual representations from expert demonstrations using a frozen vision foundation model (VFM); 2) trajectory-conditioned retrieval process, where the agent aligns its executed history with expert trajectories to identify the most distributionally consistent demonstration;  and 3) behavior retrieval stage, where the agent further refines the candidate behaviors to disambiguate visually similar states and select the next action. Fig.~\ref{fig:overview} shows an overview of ReMoBot.

\subsection{Retrieval Dataset Generation}

We first collect $n$ expert demonstrations:
\[
D_{I}^e = \{\uptau_1^e, \uptau_2^e, \dots, \uptau_n^e\},
\]

where each raw trajectory 
\[
\uptau_i^e = \{(I_{i1}^e, a_{i1}^e), (I_{i2}^e, a_{i2}^e), \dots, (I_{im_i}^e, a_{im_i}^e)\}
\]
contains RGB observations from the front-mounted camera \(I_{ij}^e \in \mathbb{R}^{H \times W \times 3}\) and the corresponding discrete actions $a_{ij}^e \in \mathbb{A}$, where $\mathbb{A}$ is a set of primitive discrete actions that may involve either the body or the arm (\eg{} body forward, arm forward, arm grasp, etc.). We then encode each observation using a frozen VFM encoder $\phi$ (Dinov2~\cite{dino_vit}), producing compact, task-relevant features. For each trajectory, we map observations into the feature space while retaining their associated actions. The resulting retrieval dataset is  
\[
D_{re}^e = \{\tau_1^e, \tau_2^e, \dots, \tau_n^e\},
\]
where each trajectory 
\[
\tau_i^e = \{(s_{i1}^e, a_{i1}^e), (s_{i2}^e, a_{i2}^e), \dots, (s_{im_i}^e, a_{im_i}^e)\}
\]
consists of encoded features $s_{ij}^e = \phi(I_{ij}^e) \in \mathbb{R}^{384}$, paired with the corresponding discrete actions $a_{ij}^e \in \mathbb{A}$, where $|\mathbb{A}|$ is the number of primitive actions. We use the upright symbol $\uptau_i^e$ for raw-observation trajectories and the italic $\tau_i^e$ for their encoded counterparts; only the latter are used hereafter. This design avoids additional training, supports generalization to unseen objects, and preserves the action-observation correspondence necessary for trajectory-conditioned retrieval.

\begin{algorithm}
\caption{ReMoBot} 
\label{alg:remobot}
\begin{algorithmic}[1]
    \State \textbf{Initialize:} 
    \State \hspace{\algorithmicindent}Given expert retrieval dataset \( D_{re}^e\);
    \State \hspace{\algorithmicindent}Online visited trajectory \(\tau^\pi \), current feature state $s_t^\pi$;
    \State \hspace{\algorithmicindent}Empty buffers \( G_{\text{sub}} \) and \( T_{\text{sub}} \); 
    
    \State \textbf{Step 1: Identify Similar States}
    \State \hspace{\algorithmicindent} Retrieve top-K  similar  expert states as \( G_{\text{sub}} \)  (Eq. \eqref{eq:re_subset}). 
    
    \State \textbf{Step 2: Extract Corresponding Trajectories}
    \State \hspace{\algorithmicindent}\textbf{for} Each \( s^e_{ij} \in G_{\text{sub}} \) \textbf{do}
        \State \hspace{\algorithmicindent}\hspace{\algorithmicindent}Retrieve the expert trajectory \( \tau^e_i \) where $s^e_{ij} \in \tau^e_i$. 
        \State \hspace{\algorithmicindent}\hspace{\algorithmicindent}Truncate \( \tau^e_i \) at timestamp $j$. 
        \State 
        \hspace{\algorithmicindent}\hspace{\algorithmicindent}Store truncated $\bar{\tau}^e_i$ in \( T_{\text{sub}} \).
    
    \State \textbf{Step 3: Evaluate Candidate Trajectories}
     \State \hspace{\algorithmicindent} Select top-L  similar expert trajectories  from $T_\text{sub}$  to form a refined set \( A_{\text{sub}} \)  (Eq.\eqref{eq:traj_sub}). 
        
    \State \textbf{Step 4: Action Selection}
    \State \hspace{\algorithmicindent}\textbf{for} Each \( \bar{\tau}^e_i \in A_{\text{sub}} \) \textbf{do}
        \State \hspace{\algorithmicindent}\hspace{\algorithmicindent} Compute the action similarity score (Eq. \eqref{eq:ham}).
    \State \hspace{\algorithmicindent}Select the $\bar{\tau}_i^e$ with highest score as \( \tau_g \).
    \State \hspace{\algorithmicindent}Execute the last action $a^e_\text{last}(\tau_g)$.
\end{algorithmic}
\end{algorithm}

\subsection {Retrieval Process}

The retrieval stage identifies relevant expert trajectories to guide action selection. Given the current observation $I^\pi_t$ and its feature $s^\pi_t = \phi(I^\pi_t)$, along with the executed trajectory history 
\[
\tau^\pi = \{(s^\pi_1,a^\pi_1), \dots, (s^\pi_t, a^\pi_t)\},
\]  

ReMoBot filters candidate trajectories from $D_{re}^e$ using two constraints: 1) state similarity and 2) trajectory similarity, as detailed in Algorithm~\ref{alg:remobot}.

\paragraph{Identify Similar States}
We first construct an initial candidate set $G_{\text{sub}}$ by retrieving the $K$ states with the smallest cosine distance in feature space:

  \begin{equation}
  \label{eq:re_subset}
    G_{\text{sub}} = \mathrm{top\text{-}K}_{s^e_{ij}\in D_{re}^e}\bigl(d_{cos}(s^\pi_t, s^e_{ij})\bigr),
\end{equation}
  \begin{equation}
   \label{eq:cossim}
    d_{cos}(s^\pi_t, s^e_{ij}) = 1 - \frac{{s^\pi_t \cdot s^e_{ij}}}{{\|s^\pi_t\| \cdot \|s^e_{ij}\|}}\;
  \end{equation}  
\paragraph{Trajectory Similarity Filtering}
Due to the partial observability inherent in ego-centric visual perception, we incorporate historical context for effective decision-making. Therefore, we prioritize within the generated state subset \( G_{\text{sub}} \) those whose associated historical trajectories closely align with the robot's actual trajectory \( \tau^\pi \). First, for each state $s^e_{ij}$ from $G_{\text{sub}}$, we retrieve the full expert trajectory $\tau^e_i$ containing $s^e_{ij}$ and truncate it at timestep $j$. The resulting sub-trajectory $\bar{\tau}^{e}_i$ is defined as:

\[ \bar{\tau}^{e}_i = \{ (s^e_{i1}, a^e_{i1}), (s^e_{i2}, a^e_{i2}), \dots, (s^e_{ij}, a^e_{ij}) \} \]

We denote this retrieved trajectory set as $T_{sub}$, which is visualized in Fig.~\ref{fig:overview} (2).

We then evaluate the similarity between the observed trajectory $\tau^\pi$ and the set of expert sub-trajectories $T_{\text{sub}}$ in order to filter out dissimilar candidates. As the similarity metric, we adopt the Wasserstein distance~\cite{wassertain}, computed solely over trajectory states. This metric captures distributional alignment and has demonstrated effectiveness in imitation learning tasks~\cite{pwil}. Formally, the Wasserstein distance between the robot’s current trajectory $\tau^\pi$ and an expert sub-trajectory $\bar{\tau}^e_i$ is given by:

\begin{equation}
 \label{eq:wasstein}
W(\tau^\pi, \bar{\tau}^e_i) = \min_{c \in C(\tau^\pi, \bar{\tau}^e_i)} \sum_{p=1}^t \sum_{q=1}^j c_{pq} \cdot d(s^\pi_p, s^e_{iq})\;
\end{equation}
where $t = |\tau^\pi|$ and $j = |\bar{\tau}^e_i|$ denote the lengths of the robot and expert sub-trajectories respectively, both treated as uniform empirical distributions. \(C(\tau^\pi, \bar{\tau}^e_i)\) is the set of all \(t \times j\) transportation matrices \(c\) satisfying the marginal conditions, with each row summing to \(\frac{1}{t}\) and each column summing to \(\frac{1}{j}\). Here, \(c_{pq}\) represents the amount of mass transported from \(s^\pi_p\) to \(s^e_{iq}\). \(d\) is an L2 distance function that evaluates the similarity between the robot's state in \(\tau^\pi\) and the expert state in the expert trajectories \(\bar{\tau}^{e}_i\), allowing us to filter out dissimilar candidates. 
Once we have estimated all the Wasserstein distance combinations between $W(\tau^\pi, \bar{\tau}^e_i)$, we keep the top-$L$ similar trajectories and form the refined skill trajectory subset $A_{\text{sub}}$:
\begin{equation}
 \label{eq:traj_sub}
 A_{\text{sub}} = \mathrm{top\text{-}L}\,_{\bar{\tau}^e_i \in T_{sub}}^{\text{min}}\bigl( W(\bar{\tau}^{e}_i, \tau^\pi)\bigr)
\end{equation}

\subsection{Action Selection}
\label{sec:motion_generation}
Given the refined trajectory candidate set $A_{\text{sub}}$, each trajectory $\bar{\tau}_i^e$ has a final state $s^e_{ij}$ that visually aligns with the robot's current observation $s_t^\pi$, and its preceding states align with the observed history. The corresponding last actions $a^e_{ij}$ form a preliminary feasible action set. However, in our ego-centric setting, multiple candidate actions may still arise due to partial observability, where the same observation can correspond to different motions. To resolve this ambiguity, we assume that expert demonstrations with similar trajectories also share similar action sequences. Guided by this intuition, ReMoBot selects actions based on action-sequence similarity. Specifically, we employ a reversed Hamming distance~\cite{hamming_dis}, which counts the number of matching positions between two action sequences, thereby favoring candidates that align more closely with the robot’s past actions. 

More specifically, denote the robot's action sequence as $\tau_A^\pi=\{a_1^\pi, a_2^\pi, ..., a_{t-1}^\pi\}$. For each $\bar{\tau}^{e}_i \in A_{sub}$, we formulate its corresponding action history sequence as $\tau_{iA}^e=\{a_{i(j-t+1)}^e, a_{i(j-t+2)}^e, ..., a_{i(j-1)}^e\}$ where $j$ is the index of the last action in $\bar{\tau}^{e}_i$. We then compute the matching score and select the trajectory $\tau_g$ with the highest matching score:
  \begin{equation}
  \label{eq:subgoal}
    \tau_g = {\arg\max}_{\bar{\tau}_i^e \in A_{\text{sub}}} d_H(\tau_{iA}^e, \tau_{A}^\pi) 
  \end{equation}
  \begin{equation}
  \label{eq:ham}
    d_H(\tau_{iA}^e, \tau_{A}^\pi)= 
    \sum_{k=1}^{t-1} \mathbf{1} \left( a_{i(j-t+k)}^{e}, a^{\pi}_k \right)\;
  \end{equation}  
where \( a^{\pi}_k \) is the action of trajectory \(  \tau^\pi \) at the \( k \)-th timestep and $a_{i(j-t+k)}^{e}$ is the action of trajectory $\tau^e_i$ at timestep $(j-t+k)$. \( \mathbf{1} \) is an indicator function that equals $1$ for identical actions and $0$ otherwise.
ReMoBot then retrieves the last action of $\tau_g$ as the feasible action $a_\text{exe} = a^e_\text{last}(\tau_g)$ to execute. This procedure ensures that $\tau_g$ is 1) visually aligned with the current observation, 2) consistent with historical visual states, and 3) temporally coherent with past actions, effectively resolving ambiguity under partial observability. Algorithm~\ref{alg:remobot} summarizes the full ReMoBot pipeline.

\begin{table}
\caption{\textbf{Dataset Details.} Average trajectory length (mean $\pm$ standard deviation) and the total number of samples.}
\label{tab:dataset}
\begin{tabular}{ccc|cc}  
\toprule
\rowcolor{white} 
\multicolumn{1}{c}{} 
& \multicolumn{2}{c}{\textbf{Simulation}}
& \multicolumn{2}{c}{\textbf{Real world}} \\
\cmidrule(lr){2-3} \cmidrule(lr){4-5}
\rowcolor{white}
 \textbf{Task} & \textbf{Length} & \textbf{Samples}  & \textbf{Length} & \textbf{Samples} \\
\midrule
\rowcolor{white}
\textbf{Uncover} &  $33.75\pm7.50$ &  675 & $30.95\pm3.46$ & 619 \\
\textbf{Cover} &  $37.1\pm6.99$ &  742 & $13.95\pm1.79$ &279  \\
\textbf{Curtain} &  $42.85\pm6.90$ &  875 & $41.1\pm5.97$ &  822 \\

\bottomrule
\end{tabular}
\figvspace{}
\end{table}

\section{Experiments}
\label{sec:experiment}
We evaluate ReMoBot on three ego-centric mobile manipulation tasks and compare it against state-of-the-art learning-based and retrieval-based baselines. Our experiments are designed to answer the following key questions:
\begin{itemize}
\item How does ReMoBot perform relative to baselines in both simulation and real-world settings?
\item Can ReMoBot generalize to variations in initial pose, object size, and material from a few demonstrations?
\item How well does ReMoBot perform under limited data?
\item What are the effects of state and trajectory constraints, and how do the hyperparameters $K$ and $L$ influence performance?   
\end{itemize}

\subsection{Mobile Manipulation Tasks}
We evaluate ReMoBot on tasks: Table Uncover, Gap Cover, and Curtain Open (Fig.~\ref{fig:task_vis}). Tasks involve complex visual observations, partial observability, and dataset imbalance, particularly in the Table Uncover and Gap Cover tasks, where the critical \textsc{GRASP} action occurs only once per trajectory. All experiments run in real time at 15 FPS on an NVIDIA RTX 3080 GPU paired with an AMD Ryzen 5000-series CPU, enabling real-time decision-making during deployment. Real-world experiments are conducted on the Boston Dynamics Spot platform, while simulation experiments are performed on the same robot in Isaac Sim. Visual features are extracted using a frozen Dinov2~\cite{dino_vit} backbone.

\noindent\textbf{Table Uncover:} In this task, the robot approaches a table and removes a covering cloth by folding and pulling it sideways. The task succeeds when the folded cloth’s edge crosses the center of the table. As this task involves a relatively short interaction horizon, it increases the risk of overfitting in learning-based methods, particularly with limited data.  

\noindent\textbf{Gap Cover:} In this task, the robot first approaches a cloth, grasps it, and covers a gap requiring coordinated body movement. The task is considered complete when one edge of the cloth fully surpasses the gap. Gap Cover involves a longer interaction horizon and more objects, reducing overfitting risk but increasing planning complexity.

\noindent\textbf{Curtain Open:} In this task, the robot approaches a curtain, pushes it aside with its arm, and then navigates its body through the opening. The task is considered successful when the curtain is sufficiently opened and the robot passes the curtain hanger. Although this task does not involve grasping, it introduces collision risk. 

\begin{figure}
    \centering
    \centerline{\includegraphics[width=0.49\textwidth]{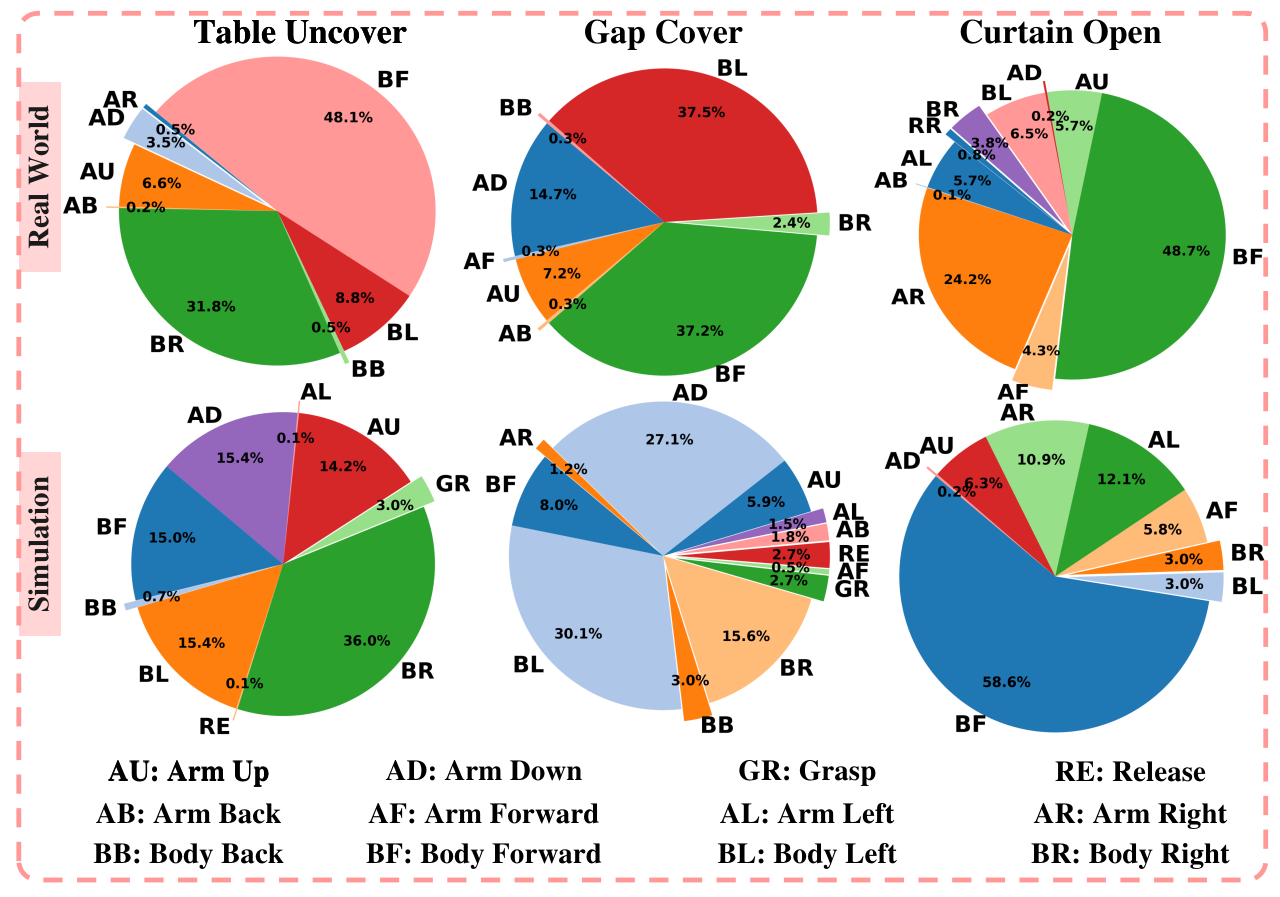}}
    \caption{\textbf{Action distribution in the dataset across tasks in both simulation and real-world settings.} Each pie chart illustrates the frequency of discrete actions, highlighting data imbalance. The rows correspond to real-world and simulation results, respectively. Tasks are ordered from left to right as Table Uncover, Gap Cover, and Curtain Open.
     \figvspace{}
    }
\label{fig:action_distribution}
\end{figure}

\noindent\textbf{Dataset Collection:} 
We first collect expert demonstrations via human teleoperation using a discrete action space composed of body and arm movement primitives. Body-level actions translate or rotate the base by a fixed step, while arm-level actions displace the end-effector along Cartesian axes. At each timestep, a single command $a^e_t$ is executed, decoupling base and arm control. Egocentric observations $I^e_{t}\in \mathbb{R}^{640 \times 480 \times 3}$ are captured from a forward-facing RGB camera mounted on the robot’s body. For the Gap Cover and Table Uncover tasks in real-world settings, a successful grasp is recorded when the end-effector contacts the deformable object. All manipulated objects (a polyester curtain and a plastic table sheet) are deformable, introducing both visual and physical challenges and increasing task complexity.

We collect 20 demonstrations per task in both simulation and real-world settings, respectively. As shown in Fig.~\ref{fig:action_distribution}, the action distribution is highly imbalanced. In the Table Uncover and Gap Cover tasks, most actions are body movements, while the critical Grasp action is extremely rare, occurring only once for each trajectory. Similarly, in the Curtain Open task, Body Forward accounts for nearly half of the dataset, whereas fine-grained arm motions are underrepresented. This imbalance poses a significant challenge for policy learning, as the agent must acquire competence in infrequent but essential actions (\eg{} Grasp) with limited data.

Table~\ref{tab:dataset} summarizes dataset statistics. The average trajectory length varies, with Curtain Open trajectories being the longest, followed by Gap Cover and Table Uncover. Real-world datasets generally contain fewer samples due to shorter trajectories, especially in the Gap Cover task. These statistics highlight both task variability and data imbalance challenges.

\begin{table}
\centering
\caption{\textbf{Baseline Comparisons in Simulation.} Success rates ( \textit{success / total trials} ) across three tasks with 40 trials each. The underlined entry denotes the best-performing method. Bold indicates methods whose performance is not significantly different from ours (Fisher’s exact test, $p \geq 0.05$).}

\label{tab:baseline_sim}
\begin{tabular}{l*6c}  
\toprule
\rowcolor{white} 
& \textbf{VINN} &\textbf{Diffusion}&\textbf{GSR}&\textbf{BAKU}  & \textbf{BC} & \textbf{Ours} \\
\midrule
\rowcolor[HTML]{F7F7F7}
\textbf{Uncover}&  0/40       &   0/40  &  6/40&  \underline{\textbf{38/40}}   & 5/40 &  \textbf{36/40}  \\
 
\textbf{Cover}     &  0/40      &   0/40  &   0/40&  \underline{\textbf{33/40}}   &   15/40 &  \textbf{31/40}\\
\rowcolor[HTML]{F7F7F7}
\textbf{Curtain}  &  0/40      &  15/40 &   12/40&  13/40 &   16/40 & \underline{\textbf{32/40}} \\

\bottomrule
\end{tabular}
\end{table}

\subsection{Baselines}
\label{sec:baseline}

We compare ReMoBot against learning-based and retrieval-based baselines. Learning-based baselines include classical Behavior Cloning (BC)~\cite{BC_survey}, the Action-Chunk Transformer (BAKU)~\cite{baku}, and Diffusion models~\cite{diffusion_policy}. Retrieval-based baselines include GSR~\cite{gsr_bc} and Visual Imitation through Nearest Neighbours (VINN)~\cite{VINN}. Except for Diffusion Policy and VINN, which take raw RGB images as input, all other methods are trained with the same embedding feature to ensure a fair comparison.

\textbf{BC}: A classical supervised learning approach~\cite{BC}, where a policy is trained to directly map observations to actions using expert demonstrations. To make the comparison fair, we train the policy to predict one-step actions.

\textbf{GSR}: A retrieval-based method~\cite{gsr_bc} that organizes the dataset into a graph and performs graph search to estimate the values of different behaviors. A retrieval procedure is then applied to identify the best behavior for each state, followed by behavior cloning to learn that behavior. For simplicity, retrieved behaviors are then cloned via a multi-layer perceptron (MLP).

\textbf{BAKU}: A transformer-based behavior cloning method that uses the history of the last $h$ observations $s_{t-h:t}$  to predict a chunk of $h$ actions via a Gaussian mixture model. Following~\cite {baku}, we set $h=5$ and execute only the first action during evaluation.

\textbf{Diffusion}: Diffuser leverages diffusion probabilistic models to generate trajectories that mimic expert behavior~\cite{ hu2024_adadiffuser}. While previous work has focused on large-scale datasets and point cloud inputs~\cite{Ze2024_3d_diffuser}, we use RGB-only observations following the Diffusion Policy framework~\cite{diffusion_policy}. 

\textbf{VINN}: VINN performs nearest neighbor search over demonstration observations to select the most similar states~\cite{VINN}, and computes an action as a Euclidean kernel-weighted average of those associated with the retrieved neighbors. We use the original VINN encoder structure, a visual representation model BYOL, with our dataset to compare against our perception pipeline.

\subsection{Baseline Comparisons}

We first set up the three tasks in the simulator to evaluate ReMoBot and all five baselines. We then train the selected methods on a real-world dataset and compare their performance with ReMoBot in real-world deployment. Notably, all methods are evaluated separately in simulation and in the real world, without any sim-to-real transfer.

\begin{figure}
    \centering
    \includegraphics[width=0.47\textwidth]{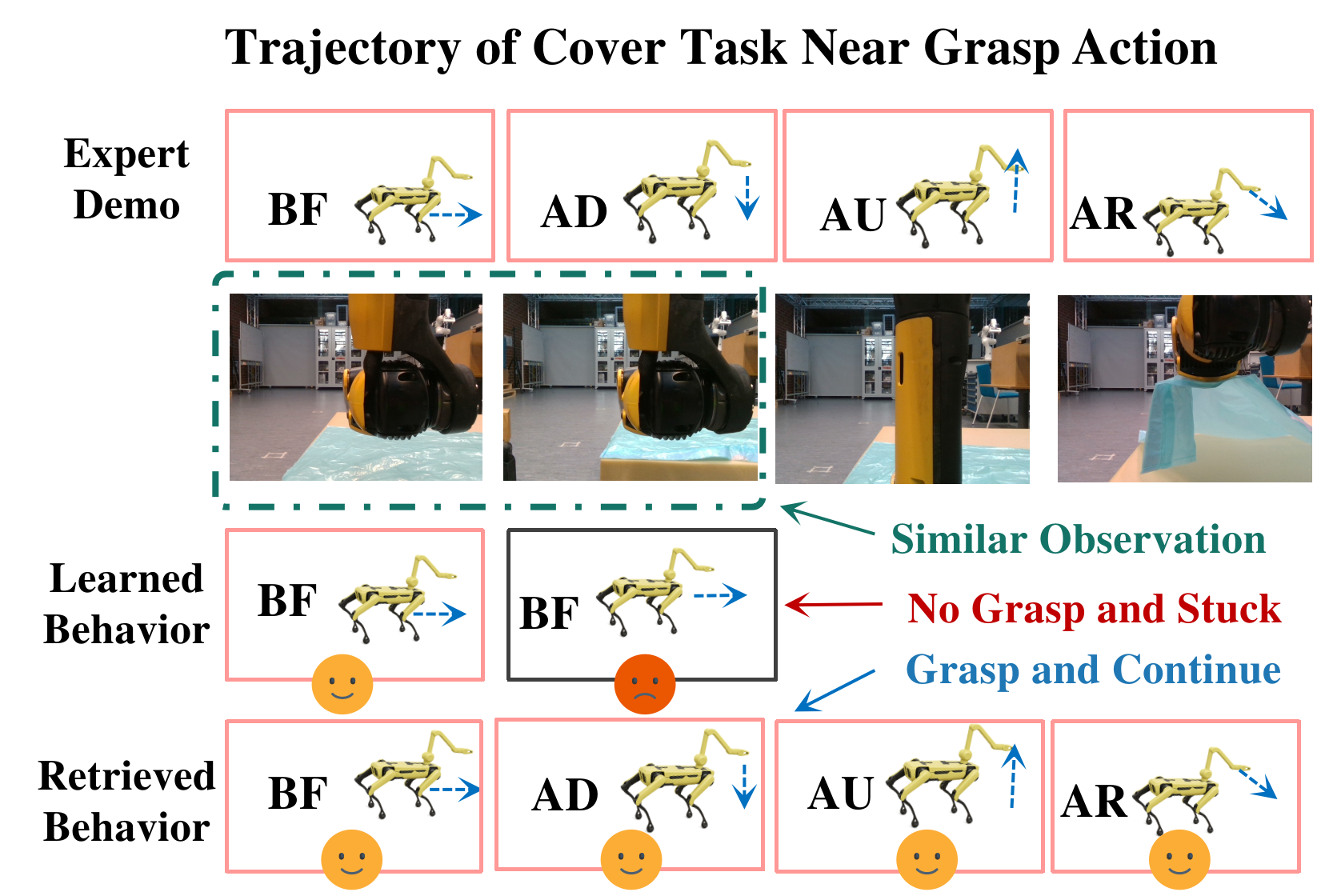}
    \caption{\textbf{Visual ambiguity near the grasp action in the Gap Cover task.} Rows show (top to bottom): expert demonstration, egocentric observations, learned behavior, and retrieved behavior (BF: move forward, AD: arm down, AU: arm up, AR: arm right). While similar pre-grasp observations cause the learned policy to fail and get stuck, ReMoBot retrieves the correct action and successfully executes the grasp.
    \figvspace{}
 }
\label{fig:failure}
\end{figure}

\begin{table*}
\centering
\caption{\textbf{Baseline Comparisons in the Real World.} Success rates (\textit{success/total trials}) over 20 trials each. The best-performing method is underlined. Bold indicates methods that are not significantly different from ours (Fisher’s exact test, $p \geq 0.05$).}
\label{tab:baseline_real}
\begin{tabular}{cccc|ccc|ccc}  
\toprule
\rowcolor{white} 
\multicolumn{1}{c}{\textbf{Tasks}} 
& \multicolumn{3}{c}{\textbf{Table Uncover}}
& \multicolumn{3}{c}{\textbf{Gap Cover}}
& \multicolumn{3}{c}{\textbf{Curtain Open}} \\
\cmidrule(lr){2-4} \cmidrule(lr){5-7} \cmidrule(lr){8-10}
\textit{\textbf{Task Stages:}}& Approach & Grasp & Uncover & Approach & Grasp & Cover & Approach & Open & Pass \\
\midrule
\rowcolor[HTML]{F7F7F7} 
\textbf{BC}       & 10/20 & 1/20  & 0/20  & 13/20 & 4/20  & 2/20  & \textbf{16/20} & 2/20  & 0/20 \\
\textbf{BAKU}     & \textbf{19/20} & 0/20  & 0/20  & 10/20 & 0/20  & 0/20  & 6/20  & 0/20  & 0/20 \\
\rowcolor[HTML]{F7F7F7} 
\textbf{ReMoBot}  & \underline{\textbf{20/20}} & \underline{15/20} & \underline{14/20} & \underline{{20/20}} & \underline{17/20}  & \underline{16/20} & \underline{\textbf{20/20}} & \underline{16/20} & \underline{9/20} \\
\bottomrule
\end{tabular}
\end{table*}

\noindent\textbf{Baseline Comparisons in Simulation:} Although all six methods generally navigate the robot close to the target (e.g., the curtain or sheet), Diffusion and VINN perform worse across most tasks (Table~\ref{tab:baseline_sim}). We hypothesize that 20 demonstrations($\sim$600 samples per task) are insufficient, compared to prior work: Diffusion~\cite{diffusion_policy} uses over 200 demonstrations, and VINN~\cite{VINN} uses 71 for training.

In addition to the dataset limitation, we identify two primary failure modes of learning-based methods. The first is the severe data imbalance across demonstrations, which biases the models toward frequently occurring actions while limiting their ability to learn rare but critical behaviors, such as the \textsc{GRASP} action in the Cover and Uncover tasks (only 2.7\% and 3.0\% of all actions, respectively). The second challenge arises from the strong visual similarity between pre- and post-grasp states, making it difficult for parametric policies to distinguish between them. To support this analysis, we manually modified BC’s output by replacing the second grasp action with an arm-up command, resulting in 33 successful trials out of 40. This intervention further validates our hypothesis that grasp misclassification is a key failure source.

In contrast, both ReMoBot and BAKU leverage historical information, enabling robust performance under limited data. Notably, BAKU performs particularly well in the Uncover and Cover tasks, where the presence of a clear bottleneck action allows its history-aware mechanism to focus on state transitions near the grasp. ReMoBot achieves competitive performance across all tasks, including the more visually complex Curtain scenario, demonstrating that retrieval-based imitation can generalize beyond specific bottleneck actions.  

Overall, these results suggest that approaches relying solely on parametric function approximation struggle to capture rare yet essential behaviors under limited data. In contrast, history-aware and retrieval-based methods are more resilient to data imbalance and observation ambiguity. These findings highlight the importance of incorporating temporal context when designing imitation learning algorithms for mobile manipulation.

Consistent with the simulation results, as shown in Tab.~\ref{tab:baseline_real}, both baselines are generally able to navigate the robot close to the target in real-world experiment. However, despite being trained on the real-world dataset with the same network structure, BAKU’s performance drops significantly. We observe the same two failure modes as in the simulation. First, in the Cover and Uncover tasks, the model often fails to issue the correct \textsc{GRASP} action, as the observations are nearly indistinguishable (Fig.~\ref{fig:failure}). As a result, the robot repeatedly executes body motions without performing the grasp, leading to task failure. This behavior further underscores the challenges posed by partial observability and highlights the advantage of ReMoBot’s retrieval mechanism in resolving action ambiguity.

\begin{table}
\centering
\caption{\textbf{Generalizability Evaluation in Real World.} Success rates (\textit{success/total trials}) of ReMoBot across variations in the object size, fabric materials, and the robot’s initial position for three tasks with 20 trials each. Underlined values indicate the best-performing condition for each task.
}
\label{tab:generalizability}
\begin{tabular}{l*3c}  
\toprule
\rowcolor{white}
  & \textbf{Table Uncover} & \textbf{Gap Cover} & \textbf{Curtain Open}  \\
\midrule
\rowcolor{white}
\textbf{Size} &  10/20 &  10/20 & 6/20  \\
\textbf{Material} &  12/20 &  11/20 &  6/20 \\
\textbf{Position} &  \underline{15/20} &  12/20 &  7/20 \\
\midrule
\textbf{Default} &  14/20 &  \underline{16/20} &  \underline{9/20} \\
\bottomrule
\end{tabular}
\end{table}

Second, we attribute the performance degradation to perception and dynamics discrepancies during real-world deployment. Variations in illumination and background clutter introduce a distribution shift between data collection and evaluation. Even under nominally identical settings, minor trajectory variations and actuator noise lead to viewpoint changes, resulting in observation differences across trials. Furthermore, collision avoidance and balance-related adjustments on the legged platform introduce additional trajectory deviations, compounding perception shifts, particularly in ego-centric settings, where observations are highly sensitive to changes in camera pose.

In contrast, ReMoBot consistently outperforms the baselines, achieving success rates of 70\% for Table Uncover, 80\% for Gap Cover, and 45\% for Curtain Open. A detailed breakdown in Table~\ref{tab:baseline_real} shows that ReMoBot is not only able to reach the target reliably (100\% success in the Approach stage across all tasks), but also executes rare but essential grasping and following actions with high accuracy. The Curtain Open task remains the most challenging due to fabric deformability, partial observability, and collision; nevertheless, ReMoBot still demonstrates improvements over the baselines.

\begin{figure}
    \centering
    \centerline{\includegraphics[width=0.46\textwidth]{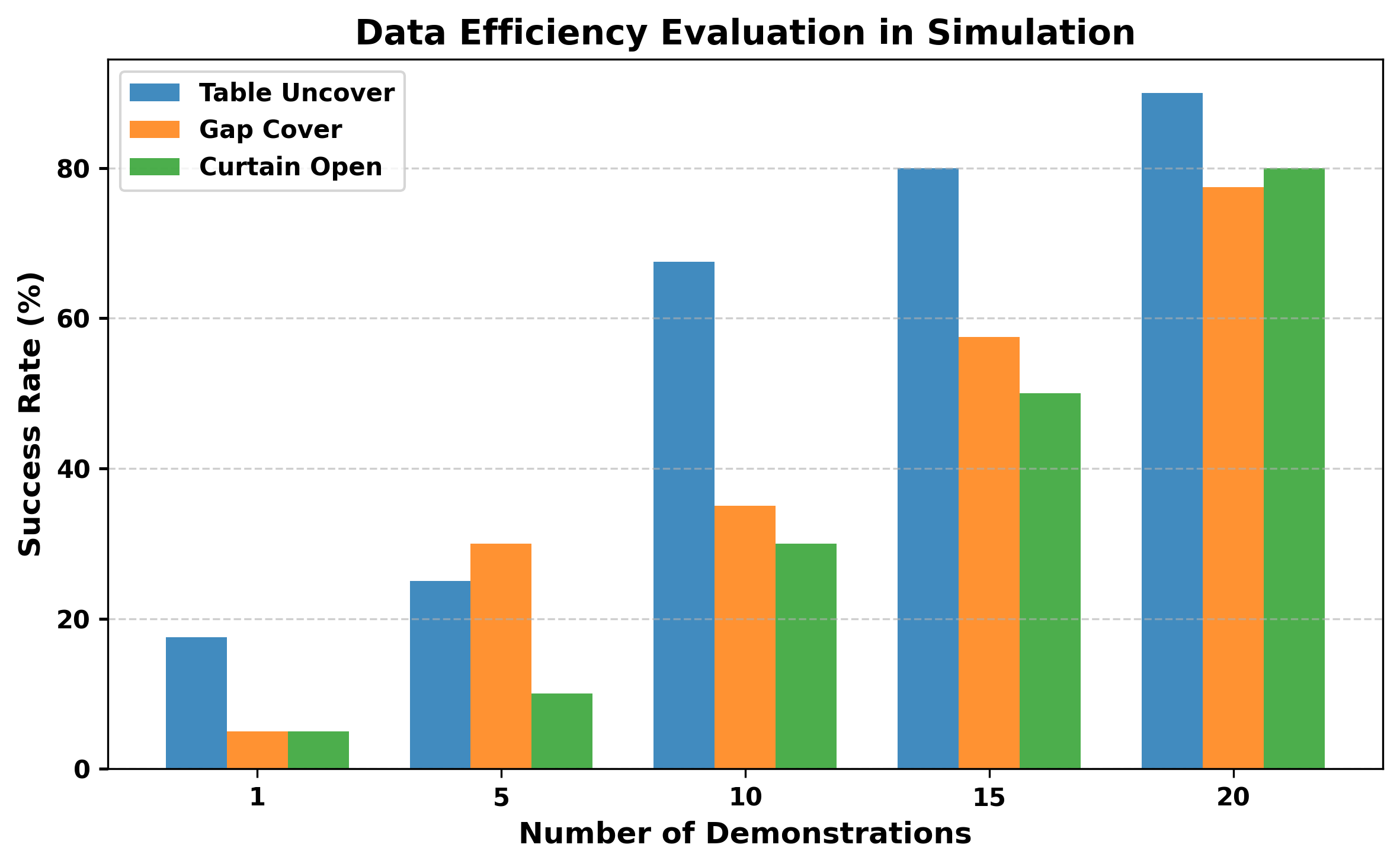}}
    \caption{\textbf{Data Efficiency Evaluation.} Success rates of ReMoBot for 40 trials across three tasks in simulation with different numbers of demonstration trajectories.
     \figvspace{}
    }
\label{fig:data_eff}
\end{figure}

\subsection{Generalizability Evaluation}
\label{sec:generalizability}
We evaluate ReMoBot in three different settings to assess its generalization capability: (1) varying object sizes, (2) different fabric materials, and (3) varying initial positions of the robot. We conduct this evaluation on the real robot with 20 demonstrations collected in the default setting. In each experiment, only one factor is varied while all other conditions remain unchanged.

\begin{itemize}
    \item \textbf{Position:} We expand the range of robot initial positions. The robot-to-curtain distance is varied from 1.5 to 1.8 meters, lateral displacement is varied up to 1 m from the curtain center, and heading angles are varied within $[-20^\circ, 20^\circ]$.

    \item \textbf{Material:} We evaluate robustness to different fabric characteristics. In the real-world environment, we evaluated performance using a blue mixed fiber (cotton and polyester) cloth and a blue plastic cover, neither of which appears in the training demonstrations. The training set contains a gray polyester curtain and a black plastic sheet.

    \item \textbf{Size:} To evaluate the influence of object sizes, we conducted tests with object sizes not used in the demonstrations. For the Curtain Open task, we tested a smaller curtain measuring 180 cm x 100 cm, where the demonstrations were performed on a 240 cm x 130 cm one. For the Cover and Uncover tasks, we employed a larger plastic cloth of 80 cm × 160 cm, compared to the 80 × 80 cm one used during the demonstration.

\end{itemize}

\begin{table}
\caption{\textbf{Metric and component ablation study in simulation.} Success rates (\textit{success/total trials}) with 40 trials each. Optimal Transport (OT) outperforms Dynamic Time Warping (DTW) for trajectory matching. ``Traj Sim.'' refers to trajectory-level visual similarity, and ``Act Sim.'' refers to action similarity scoring.}

\label{tab:metric_ablation}
\begin{tabular}{l*5c} 
\toprule
\multirow{2}{*}{\textbf{Task}} 
& \multicolumn{2}{c}{\textbf{ReMoBot}} 
& \multicolumn{2}{c}{\textbf{w/o}} 
& \multirow{2}{*}{State Only} \\
\cmidrule(lr){2-3} \cmidrule(lr){4-5}
& OT 
& DTW 
& Traj Sim. 
& Act Sim.  
&  \\

\midrule
\rowcolor{white}
\textbf{Table Uncover}    &  \underline{36/40} &  28/40 &  27/40&26/40 &30/40 \\ 
\textbf{Gap Cover}   &  \underline{31/40} &  15/40  &  21/40 &18/40&  17/40\\ 
\textbf{Curtain Open}   &  \underline{32/40} &  21/40 &  23/40 &21/40&  26/40\\ 
\bottomrule
\end{tabular}
\end{table}

Table~\ref{tab:generalizability} shows that ReMoBot maintains robust performance across diverse generalization scenarios.
Minor performance drops are primarily due to incorrect expert trajectory retrieved, when the target object is partially or entirely outside the camera’s field of view, especially with larger materials that occlude the scene. Despite these challenges, across all tasks, no condition shows statistically significant differences from the default condition (Fisher’s exact test, $p \geq 0.05$), indicating strong robustness to variations in material, size, and initial pose. 

\subsection{Data Efficiency Evaluation}
To investigate the data efficiency of ReMoBot, we conducted experiments using varying dataset sizes of 1, 5, 10, 15, and 20 demonstrations in simulation only. The evaluation environment is identical to the demonstration collection environment. Results are reported as the success rate of 40 trials. As shown in Fig.~\ref{fig:data_eff}, for simpler tasks such as Table Uncover, ReMoBot reaches around 80\% success with only 15 demonstrations, while for more challenging tasks, performance exceeds 75\% with 20 demonstrations.

\subsection{Ablation Study}

\begin{table}
\caption{\textbf{Hyperparameter Ablation in simulation.} Success rates (\textit{success/total trials}) under different hyperparameter settings over 40 trials each. The best-performing method is underlined. $K$ denotes the size of the subset $G_{\text{sub}}$, $L$ is the size of $A_{\text{sub}}$.} %
\label{tab:ablation}
\centering
\begin{tabular}{l*4c} 
\toprule
\multirow{2}{*}{\textbf{Task}} 
& K=10 &  K=10 &  K=5 & K=5 \\
 & L=5 &  L=3  &  L=5 & L=3 \\
\midrule
\rowcolor{white}
\textbf{Table Uncover}    &  \underline{36/40} &    30/40  &21/40&32/40 \\ 
\textbf{Gap Cover}    &  \underline{31/40} &    30/40  &  28/40 &  30/40 \\ 
\textbf{Curtain Open}   &  \underline{32/40} &    24/40  &  27/40 &  25/40 \\ 
\bottomrule
\end{tabular}
\end{table}

We conduct ablation studies to analyze the contribution of each component in the retrieval process, the choice of similarity metric, and the sensitivity to hyperparameters.

\noindent\textbf{Component and Metric Ablation.} Table~\ref{tab:metric_ablation} evaluates the necessity of trajectory similarity, action similarity, and the trajectory alignment metric. Removing either trajectory-level observation similarity or action similarity results in a clear performance drop across all tasks, demonstrating that both components are essential for robust decision-making. The State Only baseline also underperforms the full method, confirming that trajectory-level reasoning is necessary for handling temporal dependencies under partial observability. In addition, Optimal Transport (OT) consistently outperforms Dynamic Time Warping (DTW), indicating that OT provides more effective trajectory alignment in our setting. 

\noindent\textbf{Hyperparameter Sensitivity.} We further analyze the impact of hyperparameters in Table~\ref{tab:ablation}. Here, $K$ denotes the size of subset $G_{\text{sub}}$ in Eq.~\ref{eq:re_subset}, and $L$ denotes the size of $A_{\text{sub}}$ in Eq.~\ref{eq:traj_sub}. All experiments are conducted in simulation under the same conditions as data collection. The results show that the best performance is achieved with $K=10$ and $L=5$, which are adopted in all other experiments. 

\section{Limitations}
\label{sec:limitation}
Despite the promising results of ReMoBot, several limitations remain. First, the absence of collision-free motion planning leads to failures in the Curtain Open task, where the robot occasionally collides with the curtain hanger. Second, the retrieval module may select semantically mismatched states due to the limited power of the visual backbone and the restricted diversity of the offline dataset. This is a common limitation in imitation learning without online adaptation. Third, in the Table Uncover and Gap Cover tasks, highly similar observations can introduce local optima, suggesting the need for additional contextual signals such as a grasp flag. Finally, the decoupled control of the robot’s body and arm introduced inverse kinematics (IK)-related failures. Future work will focus on integrating collision-aware planning, enabling online adaptation and failure recovery, incorporating additional contextual cues, and exploring unified or continuous control formulations.

\section{Conclusion}
\label{sec:conclusion}
Learning mobile manipulation under partial observability from limited demonstrations remains a challenging problem. In this work, we present ReMoBot, a retrieval-based few-shot imitation learning framework that operates without additional training. By combining a visual foundation model for feature extraction with trajectory-aware retrieval and action selection, ReMoBot enables training-free imitation under partial observability. We evaluate ReMoBot on three real-world mobile manipulation tasks involving deformable objects using the Boston Dynamics Spot platform. Across all tasks, ReMoBot consistently outperforms both learning-based and retrieval-based baselines, demonstrating data efficiency and robustness to observation ambiguity. Furthermore, ReMoBot generalizes to variations in initial pose, object size, and material properties. Moving forward, future work will investigate collision-aware motion planning and online refinement mechanisms to further improve deployment robustness while maintaining the framework’s data efficiency and generalization capability.



\bibliographystyle{IEEEtran}
\bibliography{ref} 

\end{document}